\documentclass[10pt]{article}
\usepackage[T1]{fontenc}
\usepackage{hmi-preprint}
\usepackage{marvosym}
\usepackage{caption}
\usepackage{wrapfig}
\usepackage{float}
\usepackage{tikz}
\usepackage{booktabs}
\usepackage{amsmath}
\usepackage{enumitem}
\usepackage{colortbl}
\usepackage{tabularx}
\definecolor{cvprblue}{rgb}{0.21,0.49,0.74}
\hypersetup{
    colorlinks=true,
    linkcolor=cvprblue,
    citecolor=cvprblue,
    urlcolor=cvprblue,
}

\ifdefined\pdfminorversion
\fi

\preprintdate{October 2026}
\runningtitle{Interactive Navigation Beyond Reachability}
\preprintfooter{\textsuperscript{*}Equal contribution. \textsuperscript{\Letter}Corresponding author.
}

\preprintlinkset
  {https://zgca-hmi-lab.github.io/MultINav}
  {https://github.com/ZGCA-HMI-Lab/MultINav}
  {https://huggingface.co/datasets/Piqiuni/MultINav-Bench}

\preprintteaser{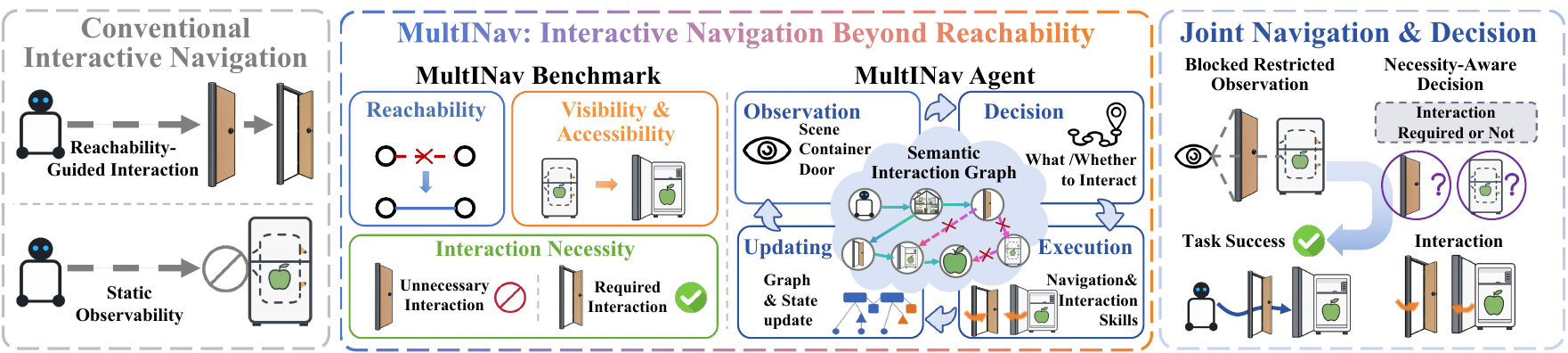}
  {\textbf{Overview of MultINav.} Channel interactions change spatial connectivity, while container interactions change target visibility. MultINav-Bench covers both effects and their sequential dependencies, with required and unnecessary episodes. MultINav-Agent uses a persistent, state-conditioned graph to connect perception, action selection, and verified state updates.}

\title{Interactive Navigation Beyond Reachability: \\
 Benchmark and Agentic Framework}

\makeatletter
\renewcommand{\@author}{%
  \textbf{Delun Li\textsuperscript{1,2,*} \hspace{0.12in}
  Xueyao Wei\textsuperscript{1,2,*} \hspace{0.12in}
  Rui Chen\textsuperscript{3} \hspace{0.12in}
  Yifei Wang\textsuperscript{3} \hspace{0.12in}\\[3pt]
  Junzhi Yu\textsuperscript{3} \hspace{0.12in}
  Wenjie Song\textsuperscript{1} \hspace{0.12in}
  Xingyu Chen\textsuperscript{2,\Letter}} \\[3pt]
  \textsuperscript{1}Beijing Institute of Technology \\[3pt]
  \textsuperscript{2}Zhongguancun Academy \\[3pt]
  \textsuperscript{3}Peking University%
}
\makeatother

\begin{document}

\maketitle

\begin{abstract}
Existing embodied navigation tasks commonly assume that traversability and object states remain fixed during an episode. Yet reaching a goal may require opening a door to access a room and then opening a container to reveal a hidden target.
We formulate interactive navigation beyond reachability, unifying channel-induced changes in spatial connectivity, container-induced changes in target visibility, and their sequential dependencies.
We introduce \textbf{MultINav-Bench}, a 3K-episode ObjectGoal benchmark in MolmoSpaces spanning channel, container, and mixed tasks, with interaction-necessity annotations covering required and unnecessary episodes. Its data generation pipeline automates task construction across multiple task types. The evaluation protocol separately measures navigation success, interaction effectiveness, and cost.
We further propose \textbf{MultINav-Agent}, a three-stage agentic framework built on a persistent, state-conditioned interaction graph. Semantic interactive perception grounds entities and interaction states; a vision-language model selects task-relevant actions based on expected changes in connectivity and target observability; and post-interaction verification updates the graph to guide subsequent navigation, exploration, or interaction.
Experiments on MultINav-Bench show that MultINav-Agent consistently outperforms adapted baselines across channel, container, and mixed tasks.
\end{abstract}

\newpage
\section{Introduction}
\label{sec:introduction}

PointGoal, ObjectGoal, and vision-and-language navigation have driven sustained progress in embodied navigation \citep{bib:anderson2018evaluation,bib:anderson2018vln,bib:chaplot2020objectnav}. These tasks test localization, mapping, search, and instruction following, but generally assume that traversability and object states remain fixed within an episode. This abstraction omits a basic property of indoor environments: interaction can change both the space a robot can traverse and the targets it can observe.

A closed door could disconnect the target room or force a long detour, while a closed refrigerator, cabinet, or drawer could conceal a nearby target. These effects can be sequential: opening a door permits entry into a room, where opening a container then reveals the target. The agent should therefore reason about how each interaction changes the environment and enables subsequent navigation, exploration, or interaction. Prior work provides foundations in semantic navigation, scene representation, and physical interaction \citep{bib:savva2019habitat,bib:kolve2017ai2thor,bib:werby2024hovsg}. For example, Interactive Gibson studies movable obstacles and the trade-off between path efficiency and physical disturbance \citep{bib:xia2020interactive}. 
Beyond free-space recovery, the task-level challenge is to determine whether an interaction restores connectivity or reveals a target, how it enables subsequent actions, and when it is unnecessary. Therefore, we are motivated to formulate the problem of \emph{multiple interactive navigation (MultINav)}, unifying \emph{channel interactions} that change spatial connectivity, \emph{container interactions} that change target visibility, and their sequential dependencies. This calls for a navigation-decision framework that jointly reasons about both types of interaction and their sequential dependencies. That is, the agent decides whether to interact, which object to select, when to act, and how to use the resulting state change to proceed toward the goal. Our motivation is shown in \autoref{fig:teaser}.

In this paper, we first instantiate the definition of MultINav as a benchmark, namely \textbf{MultINav-Bench}, a 3K-episode ObjectGoal benchmark in MolmoSpaces \citep{bib:molmospaces2026} spanning channel, container, and mixed tasks, with interaction-necessity annotations covering required and unnecessary episodes. Its data generation pipeline automates task construction across multiple task types using shared interaction-state specifications. The evaluation protocol separately measures navigation success, interaction effectiveness, and cost.
To address the MultINav problem, we propose \textbf{MultINav-Agent}, a three-stage agentic framework built on a persistent, state-conditioned interaction graph. The graph retains observed rooms and entities together with their states and relations across observations. Semantic interactive perception grounds entities and interaction states in this representation. A vision-language model uses the graph and decision history to select task-relevant actions based on expected changes in connectivity and target observability. Post-interaction verification revises affected states and relations, guiding subsequent navigation, exploration, or interaction. Our contributions are summarized as follows:

\begin{itemize}[itemsep=2pt, parsep=0pt, topsep=0pt]
    \item We formulate the problem of MultINav, which unifies channel-induced changes in spatial connectivity, container-induced changes in target visibility, and their sequential dependencies, paving the way for more comprehensive studies on interactive navigation.

    \item We introduce \textbf{MultINav-Bench}, a 3K-episode MolmoSpaces ObjectGoal benchmark spanning channel, container, and mixed tasks, with interaction-necessity annotations covering required and unnecessary episodes. As a versatile system, it provides a data generation pipeline that can automate task construction across multiple task types, and an evaluation protocol that separately measures navigation success, interaction effectiveness, and cost.
    \item We propose \textbf{MultINav-Agent}, a three-stage agentic framework built on a persistent, state-conditioned interaction graph that connects perception, action selection, and verified state updates. On MultINav-Bench, MultINav-Agent consistently outperforms adapted baselines across all task types.
\end{itemize}

\section{Related Work}
\label{sec:related-work}

\vspace{-0.2cm}
\paragraph{Embodied Navigation and Interactive Benchmarks.}
PointGoal, ObjectGoal, and vision-and-language navigation evaluate geometric, semantic, and language-specified goals under a fixed episode state \cite{bib:anderson2018evaluation,bib:batra2020objectnav,bib:anderson2018vln}. Semantic-navigation methods exploit category maps, scene graphs, and semantic priors for target search \citep{bib:chaplot2020objectnav,bib:ravichandran2022hierarchical,bib:werby2024hovsg,bib:yin2024sgnav,bib:ginting2024seek}; Habitat and AI2-THOR provide complementary environment foundations \citep{bib:savva2019habitat,bib:kolve2017ai2thor}.

\begin{table*}[t]
\centering
\scriptsize
\newcolumntype{Y}{>{\centering\arraybackslash}X}
\caption{Comparison of interaction scope and evaluated state effects.
Chan., Cont., and Obj. denote channels, containers, and manipulable objects.
Reported scale follows the respective publications.}
\label{tab:benchmark-comparison}
\begin{tabularx}{\textwidth}{@{}X Y Y Y Y X X@{}}
\toprule
\textbf{Benchmark} & \textbf{Endpoint} & \textbf{Chan.} & \textbf{Cont.} & \textbf{Obj.} & \textbf{Evaluated state effect} & \textbf{Reported scale / envs.} \\
\midrule
\multicolumn{7}{@{}l}{\textit{Non-interactive benchmarks (fixed environment state)}} \\
Habitat ObjectNav~\cite{bib:batra2020objectnav} & ObjectGoal & & & & Static & 2.8M eps. / 90 envs. \\
MolmoSpaces Navigate-to~\cite{bib:molmospaces2026} & ObjectGoal & & & & Static & 2K traj. / 679 envs. \\
R2R~\cite{bib:anderson2018vln} & InstructionGoal & & & & Static & 21.6K instr. / 90 envs. \\
VLN-CE~\cite{bib:krantz2020vlnce} & InstructionGoal & & & & Static & R2R annot. / 90 envs. \\
\midrule
\multicolumn{7}{@{}l}{\textit{Interactive benchmarks (environment-changing actions)}} \\
Interactive Gibson~\cite{bib:xia2020interactive} & PointGoal & \checkmark & & \checkmark & Free space + disturbance & 106 envs. \\
Interactive Visual Nav.~\cite{bib:zeng2021pushing} & PointGoal & & & \checkmark & Free space & 15K eps. / 120 envs. \\
CaMP~\cite{bib:wang2023camp} & PointGoal & & & \checkmark & Free space & 600K eps. / 12K envs. \\
InterNav~\cite{bib:zhou2026coins} & PointGoal & \checkmark & & \checkmark & Free space + connectivity & 150 eval / 15 envs. \\
ALFRED~\cite{bib:shridhar2020alfred} & Household & & \checkmark & \checkmark & Object state & 8K demos / 120 envs. \\
BEHAVIOR-1K~\cite{bib:li2024behavior} & Activity & & \checkmark & \checkmark & Goal predicates & 1K activities / 50 envs. \\
\textbf{MultINav-Bench (ours)} & \textbf{ObjectGoal} & \textbf{\checkmark} & \textbf{\checkmark} & & \textbf{Connectivity + target exposure} & \textbf{3K eps. / 639 envs.} \\
\bottomrule
\end{tabularx}
\vspace{-0.4cm}
\end{table*}

This fixed-state abstraction becomes limiting when an action changes the relations used for semantic navigation, such as room connectivity or target exposure. As \autoref{tab:benchmark-comparison} shows, Interactive Gibson, Interactive Visual Navigation, and CaMP retain a navigation endpoint while changing free space or path cost \citep{bib:xia2020interactive,bib:zeng2021pushing,bib:wang2023camp}; InterNav also evaluates door-mediated connectivity in PointGoal \citep{bib:zhou2026coins}. ALFRED and BEHAVIOR-1K instead use navigation to support household completion \citep{bib:shridhar2020alfred,bib:li2024behavior}. Built on MolmoSpaces \citep{bib:molmospaces2026}, MultINav-Bench retains ObjectGoal while evaluating channel reconfiguration, target-container exposure, their composition, and required/unnecessary controls.

\vspace{-0.2cm}
\paragraph{Interactive Navigation and Decision.}
NAMO introduced selecting movable obstacles to make a route feasible \citep{bib:stilman2005namo}. Interactive Gibson, Interactive Visual Navigation, and CaMP learn obstacle interactions for free-space recovery or path efficiency \citep{bib:xia2020interactive,bib:zeng2021pushing,bib:wang2023camp}; adaptive pushing, effect-oriented affordances, and joint perception-interaction-navigation extend this setting to unknown dynamics and richer action effects \citep{bib:dai2024interactive,bib:wang2024adin,bib:li2020hrl4in,bib:uppal2024spin}. Task-and-motion planning complements policy learning by grounding discrete interaction choices in geometric feasibility \citep{bib:kaelbling2011htamp,bib:toussaint2015lgp,bib:garrett2020pddlstream}. These methods establish that navigation can benefit from interaction, but generally tie the choice to obstacle clearance, path cost, or control feasibility. Language-grounded planners add language-level task decomposition, learned skill feasibility, and execution feedback \citep{bib:ichter2023saycan,bib:huang2023innermonologue,bib:singh2023progprompt,bib:huang2023voxposer}, while articulated-object systems provide reusable door-opening executors \citep{bib:li2024unidoormanip,bib:gupta2024opening}. The remaining question is whether a candidate produces a goal-relevant navigation-state change. CoINS is the closest comparator, but MultINav-Agent selects candidates by their expected graph effect and verifies that effect before replanning. It therefore distinguishes restoring connectivity from exposing a target; in MultINav-Bench, the latter becomes task-level accessibility and is assessed with required/unnecessary and distractor controls.

\vspace{-0.2cm}
\paragraph{Semantic Navigation and Structured Scene Representations.}
Dynamic and open-vocabulary scene graphs organize places, objects, relations, and traversability while supporting online construction, language queries, and task-relevant resolution \citep{bib:rosinol2020dsg,bib:hughes2022hydra,bib:gu2023conceptgraphs,bib:werby2024hovsg,bib:koch2024open3dsg}. They support graph-conditioned search and navigation \citep{bib:ravichandran2022hierarchical,bib:yin2024sgnav,bib:ginting2024seek}; graph-based and language-guided planners further connect these representations to feasible actions and replanning \citep{bib:jiao2022scenegraph,bib:ray2024tamp3dsg,bib:dai2024optimalgraph}. However, navigation graphs typically plan over relations observed during construction: even when the representation is updated online, action-induced changes in connectivity or target visibility are seldom treated as task-level state transitions that must be predicted and verified. Recent interactive graphs move toward such updates by modeling task state or articulated/container structure \citep{bib:honerkamp2024moma,bib:yan2025dovsg,bib:rotondi2025fungraph,bib:ju2025momagraph,bib:yu2026pandora,bib:buechner2026momasg}. MultINav-Agent specializes this interface to verified channel effects on connectivity and container effects on target visibility/accessibility under an ObjectGoal endpoint.

\vspace{-0.1cm}
\section{MultINav Benchmark}
\label{sec:benchmark}
\vspace{-0.1cm}

\begin{figure*}[t]
\centering
\includegraphics[width=\textwidth]{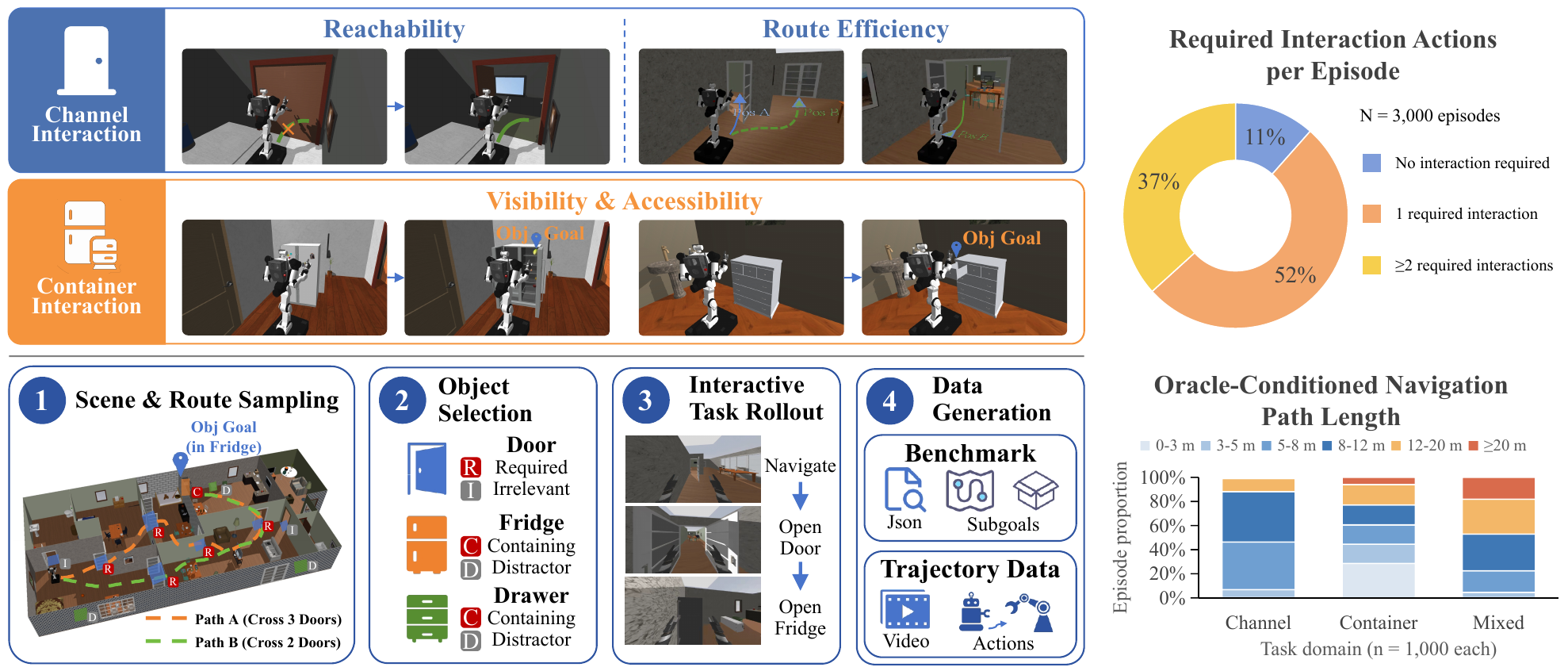}
\caption{\textbf{Construction and composition of MultINav-Bench.}
Top: channel interactions alter reachability and, in a route-shortening subset, route efficiency, whereas container interactions change target visibility and accessibility.
Bottom: benchmark construction and data collection.
Right: required-interaction counts and feasible-route lengths for the 3,000 episodes.}
\label{fig:benchmark-overview}
\vspace{-0.3cm}
\end{figure*}

We introduce \textbf{MultINav-Bench}, a diagnostic ObjectGoal benchmark for navigation under interaction-induced state changes.
Unlike static ObjectGoal navigation, an episode may require opening a passage door, exposing a target concealed inside a container, or sequentially composing both operations before navigation can succeed.
MultINav-Bench freezes 3,000 domain-balanced \emph{light} episodes in MolmoSpaces (1,000 channel / 1,000 container / 1,000 mixed across 639 houses) with explicit interaction-necessity labels, distractor objects, and oracle demonstration paths.
It evaluates whether an agent reaches the goal while purposefully deciding when and what to interact with.
\autoref{fig:benchmark-overview} summarizes the interaction effects, construction pipeline, and dataset composition.

\subsection{Task and Interaction Taxonomy}
\label{sec:task-definition}
\label{sec:interaction-taxonomy}

The embodied state is $s_t=(x_t, M_t, z_t)$, comprising the robot state $x_t$, the geometric/semantic map $M_t$, and the articulation states of interactive entities $z_t$.
Actions combine navigation $\mathcal{A}_{\mathrm{nav}}$ (updating $x_t$) and physical interaction $\mathcal{A}_{\mathrm{int}}$ (updating $z_t$, thereby modifying traversability or target visibility).
Each episode specifies the scene $\mathcal{S}_i$, initial states $(x_i^0, z_i^0)$, goal object $g_i$, annotated necessary interactions $\mathcal{I}_i^\star$, an oracle plan $\pi_i^\star$, and a ground-truth necessity label $y_i\in\{\texttt{required},\texttt{unnecessary}\}$ (formal specification in Appendix~C, ``Task Specifications and Collection Richness'').
ObjectGoal is the frozen evaluation split.

\vspace{-0.3cm}
\paragraph{Success vs.\ process.}
Terminal ObjectGoal success $S_i = \mathbb{1}[d_i^{\mathrm{pass}}\land v_i^{\mathrm{pass}}]$ requires stopping within planar distance $d_i^{\mathrm{pass}}$ with the target visible $v_i^{\mathrm{pass}}$ in the head camera.
Physical interaction enables state changes but does not establish terminal success: opening a door without locating the target yields $S_i = 0$.
ISR, IP, and normalized Total Cost separately measure required-effect progress, target-class physical effects, and navigation and interaction expenditure (Section~\ref{sec:metrics}).
The benchmark exposes two perception modes: an image-based mode using raw
visual observations, and a restricted-GT mode that supplies episode-local
identities, categories, and spatial observations for currently visible
entities. The primary comparison uses the restricted-GT mode to reduce
perception and instance-association errors as a confound while keeping
exploration, interaction selection, state updating, and terminal verification
as agent decisions. In both modes, hidden objects, simulator joint metadata,
and oracle interaction plans are withheld from the policy.

\vspace{-0.3cm}
\paragraph{Channel, container, mixed.}
\emph{Channel} interactions (hinged doors) govern spatial reachability, featuring critical closed doors (\texttt{required}), distractor doors (\texttt{unnecessary}), and composite layouts.
\emph{Container} interactions (refrigerators, cabinets, drawers) expose occluded targets, enabling the visibility required for navigation completion.
\emph{Mixed} episodes mandate sequential composition (e.g., traversing a closed door to reach a closed refrigerator).
Across the 3,000 episodes, 2,654 are labeled \texttt{required} and 346 are \texttt{unnecessary} negative controls, allowing unnecessary interaction attempts to be measured without treating all exploration as an execution error.

\vspace{-0.2cm}
\subsection{Data Generation and Collection}
\label{sec:benchmark-construction}
\label{sec:task-extensions}

Starting from ProcTHOR-10K houses and NavToObj seeds, candidates are synthesized via counterfactual reachability and visibility checks: closing a channel must sever the shortest route to the goal, while concealing a target inside a container must invalidate initial visibility until opened.
Candidates pass articulation readback, target-identity verification, and oracle rollout validation before freezing 1,000 balanced episodes per domain.

Beyond the frozen \emph{light} evaluation split designed for benchmark diagnostics, our pipeline supports a \emph{full collection mode} recording multimodal expert demonstrations.
For each verified rollout, full collection captures synchronized egocentric/allocentric visual streams, continuous base trajectories, 6D end-effector targets, and articulation states formatted into unified HDF5 records.
These trajectories provide an expert pretraining corpus for vision-language-action (VLA) models, while an accompanying interactive web inspector enables fine-grained visual auditing of base-manipulator motions.
Table~\ref{tab:benchmark-comparison} compares MultINav-Bench with prior benchmarks.

\subsection{Evaluation Metrics}
\label{sec:metrics}

We decouple terminal goal achievement from intermediate interaction efficiency using five complementary metrics:
\begin{itemize}
    \item \textbf{SR (\%)}: Terminal ObjectGoal success under standard distance and visibility thresholds.
    \item \textbf{SPL}: Success-weighted planar path efficiency relative to a feasible oracle route, reported on $[0,1]$ using the same terminal navigation-success flag as SR.
    \item \textbf{ISR (\%)}: Mean fraction of verified necessary interaction effects on required episodes; partial completion of a valid plan receives partial credit.
    \item \textbf{IP (\%)}: Category-level Interaction Precision, the episode-mean fraction of attempts producing a verified physical effect in a target interaction class. The acted-on instance need not match the annotated instance; reopening after closure can receive credit again.
    \item \textbf{Total Cost}: All-episode mean of normalized operation cost, $\min((L_i+0.3A_i+E_i)/30,1)$ on success and $1$ on failure, where $E_i$ counts attempts without a qualifying physical effect. The normalization constant $B=30$ is fixed before evaluation; it is not a step or time limit and does not stop rollouts (see Appendix~E, ``Metric Formulas'').
\end{itemize}

\begin{figure*}[!t]
\centering
\includegraphics[width=0.98\textwidth]{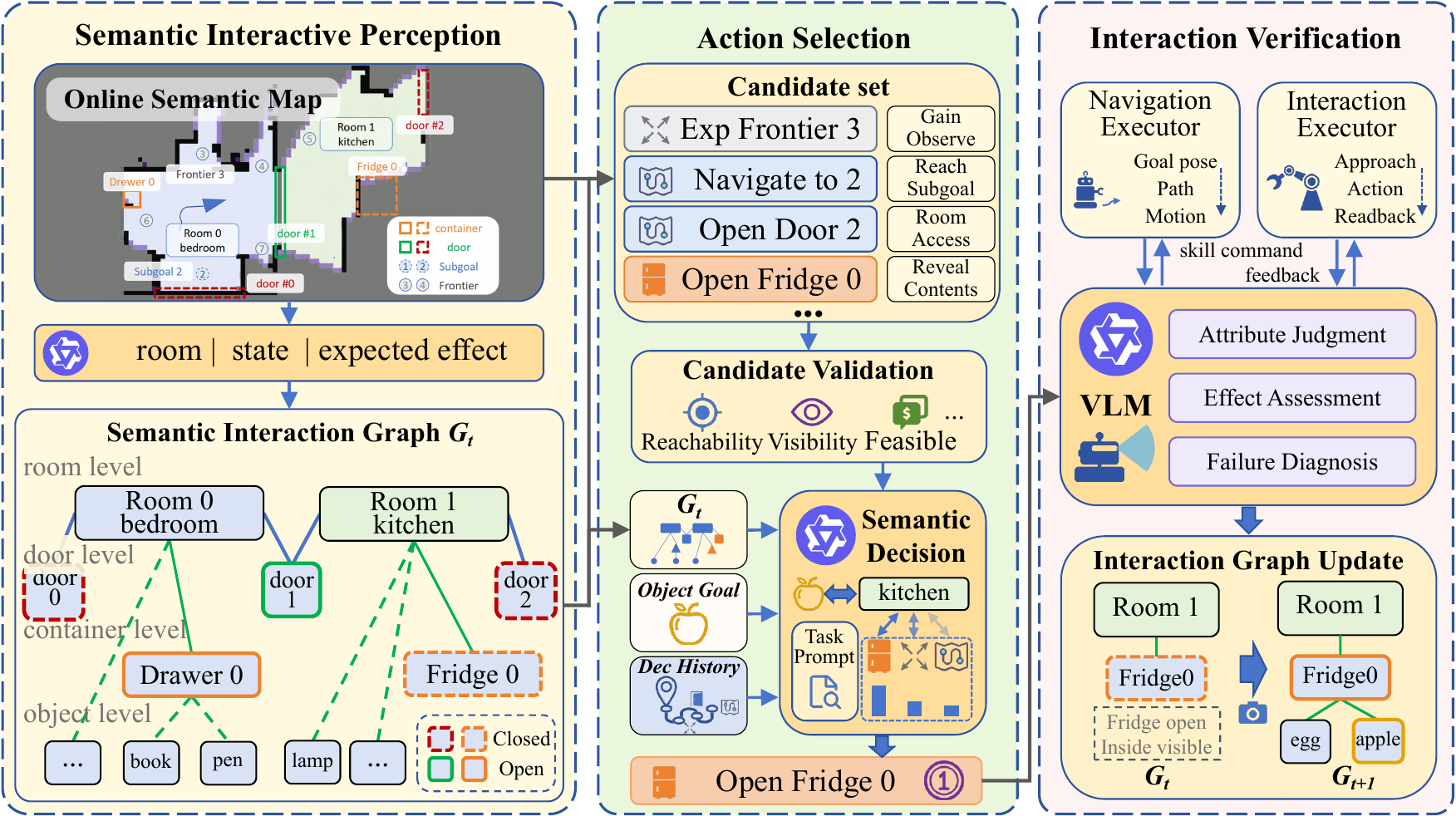}
\caption{\textbf{MultINav-Agent framework}. Perception builds a persistent, state-conditioned interaction graph; candidate plans are formed; interaction proposals are filtered for reachability, observability, and freshness; a VLM scores admissible plans by task relevance; and post-interaction verification updates the graph to guide subsequent action selection.}
\vspace{-0.3cm}
\label{fig:method-overview}
\end{figure*}

\section{MultINav-Agent Framework}
\label{sec:method}
\vspace{-0.1cm}

MultINav-Agent maintains a persistent, state-conditioned interaction graph connecting semantic perception, action selection, and post-interaction verification (\autoref{fig:method-overview}). Verified changes update the graph used to select subsequent navigation, exploration, or interaction.

\vspace{-0.2cm}
\subsection{Semantic Interactive Perception}
\label{sec:interaction-graph}

The graph $\mathcal{G}_t=(\mathcal{V}_t,\mathcal{E}_t)$ represents rooms, interactive entities, objects, and the robot through connectivity, containment, and visibility relations. Perception associates entities across observations, while visual attribute inference estimates interaction class and coarse state (open, closed, or uncertain). Interactive nodes retain candidate operations, observation confidence, and recency alongside these state estimates.

The graph is \emph{state-conditioned} because relation validity depends on the observed interaction state: a closed door blocks its mediated room connection, whereas a verified opening permits traversal. Opening a container permits inspection of its interior; target visibility is established only when the target is observed. Newly observed entities extend the graph, while previously observed entities persist beyond the current view. Their stored states and relations provide context for later decisions, with confidence and recency distinguishing historical estimates from fresh evidence (formal definitions in Appendix~D, ``Interaction Graph and State Updates'').

\vspace{-0.2cm}
\subsection{Effect-Guided Action Selection}
\label{sec:interaction-decision}

At each decision point, the graph and navigation map generate candidates for navigation, frontier exploration, and interaction. An interaction candidate specifies an entity, operation, and approach viewpoint. Eligibility checks use route feasibility, available visual evidence, and state reliability. These checks constrain the current executable candidates; remembered entities remain in the graph even when no interaction with them is currently eligible.

Each interaction candidate carries a coarse expected effect derived from its interaction class and graph relations: opening a channel may enable traversal, while opening a container may reveal its contents. This describes a possible change, without assuming an unseen target is present. A vision-language model receives the goal, relevant subgraph, candidate states and effects, route costs, and decision history. It ranks eligible actions by their expected contribution to reaching or observing the target, accounting for route prerequisites, plausible target locations, and previous attempts.

Navigation and exploration remain available when interaction offers little task progress. The selected candidate resolves to an executable goal or interaction specification; a structured reason code summarizes the model's task-level judgment.

\vspace{-0.2cm}
\subsection{Interaction Verification and Graph Updates}
\label{sec:execution-update}

For a selected interaction, the agent reaches the approach viewpoint and grounds the operable region and operation from the current view. Navigation and interaction skills execute the resulting specification. Post-interaction verification reconciles fresh visual observations with execution feedback to assess the achieved state and expected graph effect (executor details in Appendix~F, ``Additional Implementation Notes''; visual interaction prompting is summarized in Appendix~G).

Confident evidence updates the affected entity state; conflicting or occluded evidence is marked uncertain. Channel updates revise room connectivity, while container updates incorporate newly observed objects and their containment and visibility relations through perception. Failed or uncertain attempts remain in decision history to discourage immediate repetition. The updated graph regenerates candidates for subsequent navigation, exploration, or interaction.

This loop supports sequential interaction dependencies. Opening a door enables entry into a room; observations there can introduce a container candidate; verified container opening exposes objects whose observations support the next navigation decision. Each step updates the graph and available actions, composing connectivity and visibility changes during exploration.

\section{Experiments}
\label{sec:experiments}
\suppressfloats[t]
\vspace{-0.2cm}
\subsection{Experimental Setup}
\label{sec:experimental-setup}

We evaluate on MolmoSpaces NavToObj and our MultINav-Bench for non-interactive and interactive navigation, respectively. MultINav-Bench contains 3{,}000 ObjectGoal episodes spanning 639 houses in total, with balanced Channel, Container, and Mixed task splits.

\vspace{-0.2cm}
\paragraph{Evaluation scope.}
The benchmark provides both raw image-based perception and a restricted
ground-truth (RGT) perception mode. The primary comparison uses the same RGT
interface for both interactive methods: it supplies episode-local object IDs,
categories, and spatial observations for currently visible objects. This
reduces perception and instance-association errors as a confound when
evaluating interaction-aware representations and decisions; agents must still
explore, select interactions, and update their beliefs from observed outcomes.
Hidden objects, simulator joint metadata, and oracle plans are withheld.
Implementation details are provided in Appendix~A, ``Experimental Platform and Reproducibility,'' and Appendix~F, ``Additional Implementation Notes.''

\vspace{-0.2cm}
\paragraph{Interactive baseline and metrics.}
We compare MultINav-Agent (Ours) with a MolmoSpaces adaptation of MoMa-LLM~\citep{bib:honerkamp2024moma}. Both MultINav-Agent and the MoMa-LLM adaptation use Qwen3.6-35B-A3B-FP8 for high-level decisions and share the same interaction executor. We select this model to support real-time decision-making during navigation. We report SR, SPL, ISR, IP, and Cost as defined in Section~\ref{sec:metrics}, with formulas in Appendix~E, ``Metric Formulas.''

\vspace{-0.1cm}
\subsection{Main Results}
\label{sec:main-results}

\vspace{-0.1cm}
\paragraph{Non-interactive NavToObj.}
Without interaction, our method achieves 53.0\% SR on MolmoSpaces NavToObj (\autoref{tab:navtoobj-results}). The RING~\citep{bib:eftekhar2024ring} and DualVLN~\citep{bib:wei2025dualvln} leaderboard results are 48.0\% and 21.0\%; SPOC~\citep{bib:ehsani2024spoc} and VLFM~\citep{bib:yokoyama2024vlfm} score 20.15\% and 11.90\% in local evaluations. All baselines execute valid commands; lower transfer scores stem from continuous control shifts lacking sliding contact feedback.

\begin{table*}[t]
\centering
\scriptsize
\caption{\textbf{MolmoSpaces NavToObj success} ($^\dagger$leaderboard).}
\label{tab:navtoobj-results}
\newcolumntype{Y}{>{\centering\arraybackslash}X}
\begin{tabularx}{\textwidth}{@{}X Y Y Y@{}}
\toprule
\textbf{Method} & \textbf{Tasks} & \textbf{SR (\%) $\uparrow$} & \textbf{Domain} \\
\midrule
MultINav-Agent (ours) & 2{,}000 & \textbf{53.0} & ProcTHOR \\
RING$^\dagger$ & 2{,}000 & 48.0 & ProcTHOR \\
DualVLN$^\dagger$ & 2{,}000 & 21.0 & Habitat \\
SPOC & 2{,}000 & 20.15 & ProcTHOR \\
VLFM & 2{,}000 & 11.90 & Habitat \\
\bottomrule
\end{tabularx}
\end{table*}

\suppressfloats[t]
\begin{table*}[!t]
\centering
\begin{minipage}[t]{\textwidth}
\centering
\captionsetup{justification=raggedright,singlelinecheck=false,margin=10pt}
\small
\captionof{table}{\textbf{Interactive navigation results on MultINav-Bench}, grouped by method. Each method is evaluated on 3{,}000 episodes, with 1{,}000 per split. Each method's first row reports Overall, followed by task splits. Bold indicates the best value for each metric within the same split.}
\label{tab:interactive_nav_table3}
\label{tab:multinav-results}
\newcolumntype{R}{>{\raggedleft\arraybackslash}X}
\begin{tabularx}{\textwidth}{@{}X R R R R R@{}}
\toprule
& \multicolumn{2}{c}{\textbf{Navigation \& Task}} & \multicolumn{2}{c}{\textbf{Interaction}} & \multicolumn{1}{c}{\textbf{Overall}} \\
\cmidrule(lr){2-3}\cmidrule(lr){4-5}\cmidrule(lr){6-6}
\textbf{Method / Split} & \textbf{SR (\%) $\uparrow$} & \textbf{SPL $\uparrow$} & \textbf{ISR (\%) $\uparrow$} & \textbf{IP (\%) $\uparrow$} & \textbf{Cost $\downarrow$} \\
\midrule
\textbf{MoMa-LLM} & 10.53 & 0.0848 & 23.69 & 27.59 & 0.9309 \\
\hspace{1em}{\scriptsize Channel}   & 19.60 & 0.1758 & 54.28 & 40.61 & 0.8651 \\
\hspace{1em}{\scriptsize Container} & 10.30 & 0.0685 & 2.00 & 2.37 & 0.9324 \\
\hspace{1em}{\scriptsize Mixed}     & 1.70 & 0.0101 & 25.37 & 39.81 & 0.9951 \\
\midrule
\textbf{MultINav-Agent (ours)} & \textbf{41.63} & \textbf{0.2903} & \textbf{49.38} & \textbf{46.79} & \textbf{0.7743} \\
\hspace{1em}{\scriptsize Channel}   & \textbf{55.80} & \textbf{0.4567} & \textbf{60.09} & \textbf{54.29} & \textbf{0.6517} \\
\hspace{1em}{\scriptsize Container} & \textbf{45.00} & \textbf{0.3008} & \textbf{46.70} & \textbf{38.91} & \textbf{0.7220} \\
\hspace{1em}{\scriptsize Mixed}     & \textbf{24.10} & \textbf{0.1135} & \textbf{45.07} & \textbf{47.16} & \textbf{0.9493} \\
\bottomrule
\end{tabularx}

\vspace{-0.3cm}
\end{minipage}

\end{table*}

\begin{table*}[!t]
\centering
\begin{minipage}[t]{\textwidth}
\vspace{0pt}
\centering
\captionsetup{hypcap=false,justification=raggedright,singlelinecheck=false,margin=10pt}
\small
\renewcommand{\arraystretch}{1.12}
\captionof{table}{\textbf{Ablation results} on the same fixed 200-episode \textbf{Mixed subset}. Bold indicates the best value in each column.}
\label{tab:diagnostic_ablation}
\newcolumntype{R}{>{\raggedleft\arraybackslash}X}
\begin{tabularx}{\textwidth}{@{}X R R R R R@{}}
\toprule
& \multicolumn{2}{c}{\textbf{Navigation \& Task}} & \multicolumn{2}{c}{\textbf{Interaction}} & \multicolumn{1}{c}{\textbf{Overall}} \\
\cmidrule(lr){2-3}\cmidrule(lr){4-5}\cmidrule(lr){6-6}
\textbf{Method} & \textbf{SR (\%) $\uparrow$} & \textbf{SPL $\uparrow$} & \textbf{ISR (\%) $\uparrow$} & \textbf{IP (\%) $\uparrow$} & \textbf{Cost $\downarrow$} \\
\midrule
Full & \textbf{23.0} & 0.107 & 41.8 & \textbf{40.4} & 0.961 \\
Flat Object Memory & 21.0 & 0.100 & \textbf{43.8} & 26.7 & 0.963 \\
Greedy Selection & 22.0 & \textbf{0.134} & 40.9 & 36.9 & \textbf{0.933} \\
Perception-only Update & 5.5 & 0.033 & 26.6 & 36.6 & 0.988 \\
\bottomrule
\end{tabularx}
\end{minipage}
\end{table*}

\vspace{-0.2cm}
\paragraph{Interactive benchmark.}
\autoref{tab:interactive_nav_table3} shows that MultINav-Agent achieves better aggregate values on all five metrics than the MolmoSpaces adaptation of MoMa-LLM. Ours completes 1{,}249/3{,}000 episodes, reaching 41.63\% SR versus 10.53\% (+31.10 percentage points), with SPL increasing from 0.0848 to 0.2903. ISR increases from 23.69\% to 49.38\%, IP from 27.59\% to 46.79\%, and normalized Cost decreases from 0.9309 to 0.7743.

\vspace{-0.2cm}
\paragraph{Comparison across interaction types.}
Our MultINav-Agent achieves 55.80\%, 45.00\%, and 24.10\% SR on Channel, Container, and Mixed tasks, exceeding MoMa-LLM by 36.20, 34.70, and 22.40 percentage points, respectively. Channel ISR differs by only 5.81 points (60.09\% vs. 54.28\%), despite the much larger SR gap, underscoring the distinction between achieving an interaction effect and reaching the final goal. Container tasks show the largest ISR gap (46.70\% vs. 2.00\%), alongside higher target-class IP (38.91\% vs. 2.37\%). Mixed tasks remain the hardest for both methods: MoMa-LLM reaches 25.37\% ISR but only 1.70\% SR, while our method reaches 45.07\% ISR and 24.10\% SR. Partial interaction progress therefore remains insufficient for completing combined channel-container tasks.

\vspace{-0.2cm}
\paragraph{Interaction-to-goal gap.}
ISR averages partial completion over required episodes, so it need not track terminal SR. For Ours, 393/654 required Channel episodes complete all necessary effects, and 297/393 also reach the goal (75.6\%). Container and Mixed complete-effect counts are 466/1{,}000 and 247/1{,}000, of which 435/466 (93.3\%) and 223/247 (90.3\%) reach the goal. Completing the required interaction chain is thus a prominent difficulty for Ours on Container and Mixed tasks; Channel retains a larger navigation gap after the effects are achieved. Mixed ISR of 45.07\% includes partial credit, whereas only 24.70\% of episodes complete all necessary effects.

\paragraph{Efficiency and failure patterns.}
Ours has higher SPL and lower normalized Cost in every split. Nevertheless, both methods incur near-ceiling Cost on Mixed tasks (0.9493 for Ours and 0.9951 for MoMa-LLM). Since Cost is one for navigation failures, the 759 Mixed failures for Ours contribute 0.759 to its mean Cost. Similarly, SPL is gated by success. These aggregate improvements therefore reflect both completion and trajectory efficiency, rather than establishing shorter paths or cheaper execution conditional on success. Ours records exploration stalls on 516 Mixed episodes, compared with 304 Container and 158 Channel episodes, motivating further analysis of progress across interacting subgoals. These terminal labels do not establish a causal failure mechanism.

\begin{figure*}[!t]

\centering
\includegraphics[width=0.98\textwidth]{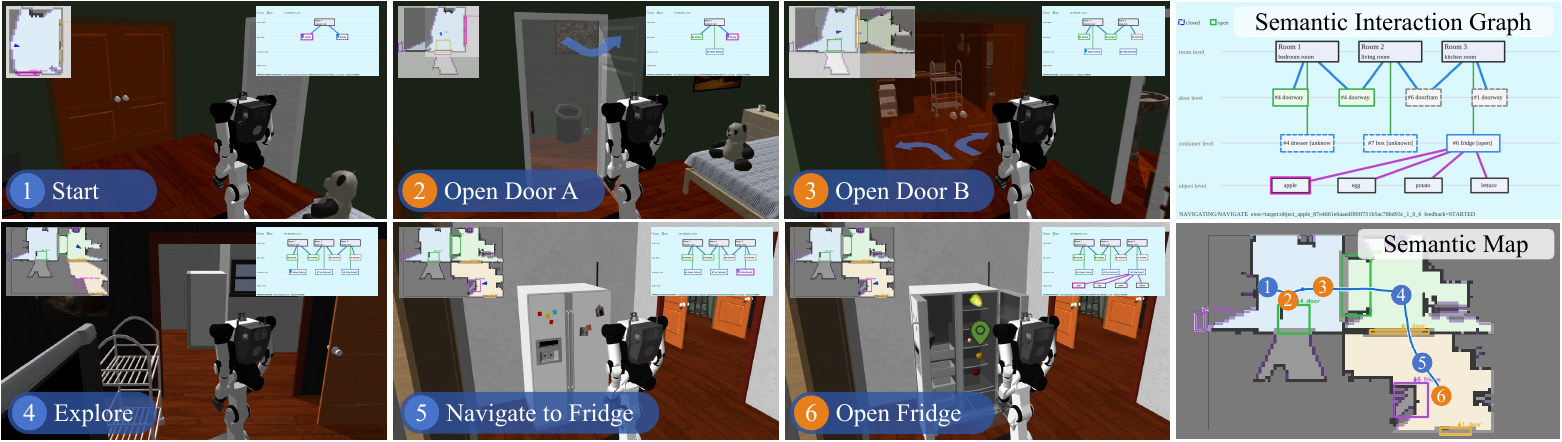}
\caption{\textbf{Sequential interactions and graph updates.}
The robot opens two doors to reach a refrigerator, then opens it to reveal the target.
Insets show evolving semantic maps and interaction graphs; right: final graph and trajectory
(blue: navigation/exploration; orange: interactions).}
\label{fig:case-study}
\par\vspace{0.6em}
\input{Figures/fig_real_world}
\vspace{-0.3cm}
\end{figure*}

\subsection{Ablation Study}
\label{sec:ablation}

All four variants use the same fixed 200-episode subset of the 1{,}000 Mixed tasks, spanning 165 houses. We sample proportionally without replacement by container class and oracle-restored reference path length, using benchmark metadata only for sampling. The episode list is frozen before evaluation without using policy outcomes; construction details appear in Appendix~B.5, ``Ablation Subset Construction.''

Table~\ref{tab:diagnostic_ablation} examines interaction memory, action selection, and state updating. Full achieves the highest SR (23.0\%) and IP (40.4\%), but the variants reveal different trade-offs rather than uniform gains across all metrics.

\vspace{-0.2cm}
\paragraph{Interaction memory.}
Flat Object Memory reduces IP from 40.4\% to 26.7\% and SR from 23.0\% to 21.0\%, despite increasing ISR from 41.8\% to 43.8\%. The graph-based memory therefore shows its clearest advantage in interaction precision, not required-effect completion. This pattern is consistent with relational context helping select effective target-class interactions, while the higher ISR of the flat variant shows that partial progress alone need not translate into goal completion.

\vspace{-0.2cm}
\paragraph{Action selection.}
Greedy Selection attains higher SPL (0.134 vs. 0.107) and lower Cost (0.933~vs. 0.961), with slightly lower SR (22.0\%) and IP (36.9\%). Full completes only two additional episodes out of 200, so these results do not establish a decisive success advantage for semantic selection. Instead, they expose a completion-efficiency trade-off: goal-conditioned selection improves the observed SR and IP, whereas the greedy variant performs better on aggregate efficiency. Since SPL and Cost also depend on success, these scores do not isolate efficiency conditional on completion.

\paragraph{State updating.}
Perception-only Update produces the largest SR decrease, from 23.0\% to 5.5\% (17.5 percentage points), alongside lower ISR (26.6\%), lower SPL (0.033), and higher Cost (0.988). IP remains 36.6\%, indicating that some interactions still achieve target-class effects without completing the task. Among the tested variants, this is the strongest evidence for retaining execution-informed graph updates when coordinating channel and container interactions. The smaller SR differences for memory and selection should be interpreted cautiously on this fixed subset, rather than as established statistically significant gains.

\subsection{Case Studies}
\vspace{-0.2cm}
\label{sec:case-studies}

\autoref{fig:case-study} illustrates an episode combining channel and container interactions: the robot opens two doors to expand reachability, navigates to the refrigerator, and opens it to expose the target. Per-step insets show how verified state updates guide continuous replanning.

\vspace{-0.3cm}
\subsection{Real-World Demonstration}
\label{sec:real-world}

\vspace{-0.2cm}
\paragraph{Setup.}
We conduct the real-world demonstration on a Go2 quadruped in an indoor environment (\autoref{fig:real-world}). Door and refrigerator opening are performed with human assistance. Egocentric observations, intermediate interaction graphs, and the semantic map record the evolving scene representation throughout the rollout.

\vspace{-0.2cm}
\paragraph{Task and rollout.}
The task is to find an apple. Starting near a closed door, the robot encounters a channel interaction: the door is opened with human assistance, enabling access to the adjacent space. The robot then explores the newly accessible area, extending its semantic map and interaction graph as rooms and objects are observed. It subsequently navigates to the refrigerator, which is opened with human assistance so that the robot can inspect its contents for the target apple. This sequence links two distinct state changes within a single object-search task: door opening changes spatial connectivity, while refrigerator opening exposes previously occluded contents. The intermediate graphs in \autoref{fig:real-world} show the corresponding evolution of room, door, container, and object relations.

\section{Conclusion}
\label{sec:conclusion}
\vspace{-0.2cm}
We formulated MultINav, unifying channel and container interactions within ObjectGoal navigation, and introduced MultINav-Bench (3{,}000 episodes) together with MultINav-Agent, whose persistent interaction graph with verified state updates achieves 41.63\% SR versus 10.53\% for the adapted baseline and transfers to a real-world Go2 demonstration.
Extending to broader interaction types, learned perception, and end-to-end policy training are promising future directions.

\bibliographystyle{unsrtnat}
\bibliography{main}

\clearpage
\appendix

\section{Experimental Platform and Reproducibility}
\label{app:experimental-platform}

\paragraph{Runtime environment.}
The simulation experiments use NVIDIA A100 GPUs with 80\,GB of memory.
The real-world demonstration uses one NVIDIA GeForce RTX 4090 with 48\,GB
of memory. The simulation software stack uses Python 3.11.15,
MuJoCo 3.4.0, and PyTorch 2.7.1.

\paragraph{Evaluation configuration.}
The simulation uses the MolmoSpaces ObjectGoal protocol, with terminal success
requiring both goal distance and head-camera visibility. Both compared methods
receive RGB-D observations and the same restricted records for currently visible
objects, and share the interaction executor. The reported operation-cost metric
uses $\lambda=0.3$, $\mu=1$, and $B=30$ (Section~\ref{app:metrics}); $B$ is
a normalization constant rather than a rollout limit.

\section{MultINav-Bench Distribution Details}
\label{app:benchmark-dist}

This appendix expands the light ObjectGoal split summarized in the main paper
(3{,}000 episodes; 1{,}000 channel / 1{,}000 container / 1{,}000 mixed; 639 houses).
All counts are taken from the frozen balanced manifest and quality reports.
Primary claims in the main paper do not depend on reading this appendix.

\subsection{Domain, Recipe, and Necessity}
\label{app:domain-recipe}

\begin{table}[H]
\centering
\scriptsize
\setlength{\tabcolsep}{3pt}
\caption{Domain recipes and necessity labels in MultINav-Bench (light).}
\label{tab:app-recipes}
\begin{tabular}{lrr}
\toprule
\textbf{Split / recipe} & \textbf{$N$} & \textbf{Notes} \\
\midrule
Channel total & 1000 & 253 houses \\
\quad single-path door closed & 366 & required \\
\quad distractor doors closed & 346 & unnecessary \\
\quad critical+distractor & 288 & required \\
Container total & 1000 & 569 houses \\
Mixed total & 1000 & 369 houses \\
\midrule
Required (all) & 2654 & \\
Unnecessary (all) & 346 & channel only \\
\bottomrule
\end{tabular}
\end{table}

Container and mixed episodes are labeled required under the construction-time
initial closed state and its validated reference plan. This label describes
the intended interaction dependency; terminal success is assessed separately
from completion of the annotated effects.
Unnecessary episodes are exclusively the channel distractor-door recipe.
Ambient articulated objects outside the listed necessary set are not separate
episode families; spurious interactions are penalized by Interaction Precision
and Total Cost in the main metrics.

\subsection{Path Length and Interaction Count}
\label{app:path-stats}

\begin{table}[H]
\centering
\scriptsize
\setlength{\tabcolsep}{3pt}
\caption{GT path length (m) and mean required-interaction count
from the light manifest.}
\label{tab:app-path}
\begin{tabular}{lrrrr}
\toprule
\textbf{Domain} & \textbf{$N$} & \textbf{Path med} & \textbf{Path mean} & \textbf{Int.\ mean} \\
\midrule
Channel & 1000 & 8.12 & 8.71 & 0.65 \\
Container & 1000 & 5.93 & 7.99 & 1.10 \\
Mixed & 1000 & 11.50 & 13.58 & 2.01 \\
All & 3000 & 8.69 & 10.09 & 1.25 \\
\bottomrule
\end{tabular}
\end{table}

Mixed episodes are longer on average because they chain a channel interaction
and a container interaction. Channel interaction counts below 1 on average
reflect the unnecessary distractor subset (zero required interactions).

\subsection{Target-Category Skew}
\label{app:category-skew}

ObjectGoal targets are not uniform over categories.
The top categories in the balanced 3K split include egg (439; 14.6\%),
tomato (199; 6.6\%), apple (193; 6.4\%), lettuce (168; 5.6\%), and potato (150; 5.0\%).
Long-tail categories (e.g., plate, soapbar, spraybottle) appear only a few times.
We report this skew for transparency; main tables use split macro averages so that
no single category dominates the headline comparison by construction, though residual
imbalance remains.

Container interaction hardware types in the 3K light set are dominated by
hinged container doors and sliding drawers (plus channel hinged doors in
channel/mixed episodes). Exact type histograms are released with the manifest
audit JSON.

\subsection{Per-House Multiplicity}
\label{app:per-house}

Episodes span 639 unique houses.
Per-domain house counts are 253 (channel), 569 (container), and 369 (mixed).
Balancing applies soft per-house caps during collection; the frozen set still
admits multiple episodes per house (empirical maximum 16 episodes in one house;
high tail near 11 at the top 5\% of houses by count).
Train/eval separation is enforced at the house level in the collector configuration
so that the same house does not appear in both partitions of a formal split.

\subsection{Ablation Subset Construction}
\label{app:ablation_sampling}

We constructed a fixed 200-episode subset from the 1,000 mixed
episodes in the benchmark, covering 165 houses. Every episode
requires both channel and container interactions. Sampling preserves
variation in container class and reference path length without using
policy outcomes. Full and all three ablated variants use the same
fixed subset.
The subset is sampled without replacement from benchmark indices
2000--2999 with pseudorandom seed \texttt{20260926}. For auditability, the
source benchmark and frozen selection-file SHA256 values are:
\begin{quote}\small
Benchmark: \texttt{4021ca2bebd9c875ccc4df70c746d9ed}\\
\texttt{7f2376d13247fa1b7e98f2e9690b219f}\\[2pt]
Selection file: \texttt{30264942a1caf203711b55442352e2a5}\\
\texttt{b0e33ea701f05a3efffb03cca136c1d3}
\end{quote}

\subsection{Scoring Eligibility and Quality Gates}
\label{app:scoring-eligible}

Before formal aggregation, the evaluator applies a runtime goal-consistency gate:
frozen oracle terminal waypoints must remain consistent with the live pose of the
selected instance.
Traces that fail are retained for diagnosis but marked scoring-ineligible and
excluded from primary denominators.
Schema audit on the balanced 3K reports 3{,}000/3{,}000 valid episodes,
0 duplicate case IDs, and 0 missing comparable GT path lengths in the offline
health report; runtime eligibility is a separate online check.

\subsection{Domain-Specific Collection Gates}
\label{app:collection-gates}

All builders emit the shared episode schema, but use
domain-specific acceptance tests.
For \emph{channel} episodes, the collector starts from a fixed Navigate-to
scene-start-goal tuple, identifies doors associated with candidate room routes,
and validates closed- and open-state path feasibility.
Required episodes must recover a feasible goal route after the annotated door
effect; unnecessary episodes retain a feasible route while closed distractor
doors remain present.
A small route-shortening pilot additionally retains cases in which both states
are feasible but opening reduces route length; it is diagnostic rather than a
balanced MultINav-Bench stratum.

For \emph{container} episodes, the collector binds a specific rigid target to a
moving fridge, cabinet, or drawer compartment, closes the relevant articulation,
and verifies the target-compartment association.
It then validates an interaction pose, the opening effect, and a terminal pose
that satisfies the same distance-and-visibility predicate as Navigate-to.
Other articulated objects remain in the scene as ambient distractors.
For \emph{mixed} episodes, the full ordered door-container rollout must pass;
the post-door state alone is not terminal, and omitting either annotated effect
must break the validated chain.
Across domains, failed state readback, inconsistent target identity, oracle
rollout failure, duplicates, and split leakage cause rejection.

\section{Task Specifications and Collection Richness}
\label{app:pg-ig}
\label{app:collection-axes}

The generator varies task specification and record richness independently.
Only the ObjectGoal-light cell is frozen as MultINav-Bench; the other cells reuse the
same interaction entities, object roles, expected effects, and oracle transitions.

\begin{table}[H]
\centering
\scriptsize
\setlength{\tabcolsep}{3pt}
\caption{Three task specifications supported by the shared collector.}
\label{tab:app-task-forms}
\begin{tabular}{lp{0.68\columnwidth}}
\toprule
\textbf{Task form} & \textbf{Stored goal specification} \\
\midrule
ObjectGoal & Target category/instance, distance and head-camera visibility success criteria; primary MultINav-Bench split. \\
PointGoal & Grounded planar goal (optional yaw) with interaction-aware route feasibility and path-delta annotations. \\
InstructionGoal & Text grounded to oracle steps and entity IDs, with hidden/partial/explicit interaction disclosure. \\
\bottomrule
\end{tabular}
\end{table}

\paragraph{Light evaluation records.}
Light records formally specify each benchmark episode as
\begin{equation}
e_i = (\mathcal{S}_i, x_i^0, g_i, z_i^0, \mathcal{I}_i^\star, \pi_i^\star, y_i),
\label{eq:app-episode-tuple}
\end{equation}
storing scene identifier $\mathcal{S}_i$, initial robot pose $x_i^0$, goal specification $g_i$,
initial articulation states $z_i^0$, annotated task-relevant interaction set $\mathcal{I}_i^\star$,
oracle navigation-interaction plan $\pi_i^\star$, ground-truth necessity label $y_i\in\{\texttt{required},\texttt{unnecessary}\}$,
and construction quality evidence.
They are sufficient to reconstruct and score an episode without storing a policy rollout.

\paragraph{Full trajectory records.}
Full records contain the light schema plus timestamped synchronized visual observations (head and external views),
executed base actions, 6D end-effector target poses, and rendered video from a chosen
navigation/interaction policy.
They may support learning, behavior analysis, or failure diagnosis, but are not
required to score MultINav-Bench and are not assumed to train the evaluated policy.

\paragraph{PointGoal.}
Candidates are sampled on the robot-inflated occupancy map with start-goal
distance bounds and require an open-state GT path.
Comparing closed- versus open-state paths yields required (no closed path),
beneficial (closed path longer by a margin), or unnecessary labels.
Beneficial strata can be denser than in the MultINav-Bench ObjectGoal split because path-delta
filtering is built into the sampler.

\paragraph{InstructionGoal.}
Layered generation over a grounded oracle plan supports rule, LLM, VLM, and
graph-grounded text, with stored entity IDs, plan-step indices, disclosure level
(hidden / partial / explicit), and optional VLM evidence frames.
Instructions are auditable annotations rather than free text detached from the
interaction graph.

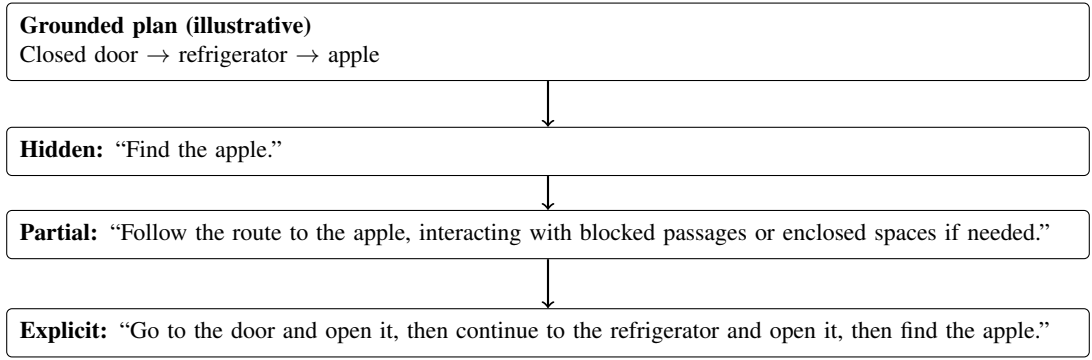
\begin{figure}[H]
\centering
\begin{tikzpicture}[
  every node/.style={font=\small, align=left},
  box/.style={draw, rounded corners=2pt, inner sep=5pt,
    text width=.80\linewidth},
  link/.style={->, thick}
]
\node[box] (plan) at (0,0)
  {\textbf{Grounded plan (illustrative)}\\Closed door $\rightarrow$ refrigerator $\rightarrow$ apple};
\node[box] (hidden) at (0,-1.45)
  {\textbf{Hidden:} ``Find the apple.''};
\node[box] (partial) at (0,-2.55)
  {\textbf{Partial:} ``Follow the route to the apple, interacting with blocked passages or enclosed spaces if needed.''};
\node[box] (explicit) at (0,-3.85)
  {\textbf{Explicit:} ``Go to the door and open it, then continue to the refrigerator and open it, then find the apple.''};
\draw[link] (plan.south) -- (hidden.north);
\draw[link] (hidden.south) -- (partial.north);
\draw[link] (partial.south) -- (explicit.north);
\end{tikzpicture}
\caption{Illustrative InstructionGoal disclosure levels generated from one
grounded interaction plan. These are task-specification examples, not results
from a frozen InstructionGoal evaluation split.}
\label{fig:app-instruction-disclosure}
\end{figure}

\section{Interaction Graph and State Updates}
\label{app:graph-updates}

Let $\mathcal{G}_t=(\mathcal{V}_t,\mathcal{E}_t(z_t))$ be the interaction graph.
Its nodes encompass rooms $\mathcal{R}$, interactive entities $\mathcal{U}$, and
candidate objects $\mathcal{O}$; each entity $u\in\mathcal{U}$ maintains a state
$z_t(u)\in\{\texttt{open},\texttt{closed},\texttt{uncertain}\}$.
The mediated room-edge weight and container visibility relation are
\begin{equation}
\begin{aligned}
w_t(R_i, R_j \mid u_{\text{door}}) &=
\begin{cases}
d(R_i, R_j) & \text{if } z_t(u_{\text{door}}) = \texttt{open}, \\
\infty & \text{if } z_t(u_{\text{door}}) \in \{\texttt{closed},\texttt{uncertain}\},
\end{cases} \\
\text{Vis}_t(o \mid u_{\text{cont}}) &= \mathbb{1}[z_t(u_{\text{cont}}) = \texttt{open} \land o \in \text{FOV}_t],
\end{aligned}
\label{eq:graph-state-condition}
\end{equation}
where $w_t$ governs door-mediated traversal and $\text{Vis}_t$ governs exposure
of an object inside a container. An uncertain door is conservatively blocked
until its state is verified; an uncertain container does not establish visibility.
The graph is persistent across observations: an entity that leaves the current
field of view is retained with its last state, confidence, and observation time.
A later observation is associated with an existing entity when the instance
evidence is compatible; otherwise it creates a new candidate entity. Fresh
evidence replaces stale attributes, while contradictory evidence reduces
confidence and can move the state to \texttt{uncertain}. Absence from a new
view is therefore not treated as disappearance, closure, or negative evidence.
Candidate edges are regenerated from the current map and graph state, while
the underlying entity history remains available to action selection.

For an interaction targeting $u^\star$ with requested state $z_{\text{target}}$
and post-interaction observation $o_{t+1}$, the verifier updates the entity by
\begin{equation}
z_{t+1}(u^\star) =
\begin{cases}
z_{\text{target}} & \text{if post-action evidence verifies the requested state}, \\
\texttt{uncertain} & \text{if evidence is conflicting, partial, or occluded}, \\
z_t(u^\star) & \text{if execution fails or times out without observed change},
\end{cases}
\label{eq:entity-state-update}
\end{equation}
The verifier requires confidence at least $\tau_{\text{verify}}$ that the
observed state matches $z_{\text{target}}$; confidence in a different state
cannot justify writing the requested state. A verified effect takes precedence
over a failed macro acknowledgement, while partial effects retain uncertainty
until the resulting state can be established. The graph update is
\begin{equation}
\mathcal{G}_{t+1} = \Phi(\mathcal{G}_t, z_{t+1}(u^\star), o_{t+1}) = \bigl(\mathcal{V}_t \cup \mathcal{V}_{\text{new}}(o_{t+1}),\; \mathcal{E}_t(z_{t+1}) \cup \mathcal{E}_{\text{new}}(o_{t+1})\bigr),
\label{eq:graph-update}
\end{equation}
reweighting door-mediated room edges or adding newly exposed objects and their
containment and visibility relations after a container is opened.

\section{Metric Formulas}
\label{app:metrics}

Let $N$ be the number of scoring-eligible episodes in the aggregation scope,
$S_i=\mathbb{1}[d_i^{\mathrm{pass}}\land v_i^{\mathrm{pass}}]$ the verified
ObjectGoal success indicator, $L_i$ the executed planar base path length,
$A_i$ all interaction attempts, and $E_i$ attempts without any qualifying
physical effect, counted at most once per attempt. A verified partial macro
effect is not an error, even if the macro reports failure; an effect in a
non-target class is likewise not automatically an error. For the restricted
ROS policy, success additionally requires a completion claim verified against
published target evidence and the benchmark distance criterion.

\paragraph{SR.}
$\mathrm{SR}=\frac{1}{N}\sum_i S_i$.
Interaction completion is neither necessary nor sufficient in this definition.
An episode can satisfy the terminal goal through an alternative observed route
or a visibility change without completing every annotated reference effect.
Such cases retain their construction-time necessity label and are scored by
the terminal goal predicate rather than being retroactively relabeled.

\paragraph{SPL.}
\[
\mathrm{SPL}_i
=S_i\frac{L_i^{\mathrm{ref}}}
{\max(L_i^{\mathrm{ref}},L_i)},
\qquad
\mathrm{SPL}=\frac{1}{N}\sum_i\mathrm{SPL}_i,
\]
where $L_i^{\mathrm{ref}}$ is a feasible reference that permits necessary
interactions (oracle plan), not the initial static shortest path.
Failed episodes contribute zero; successful zero-reference, zero-motion
episodes contribute one. Missing or invalid reference lengths make the
aggregate SPL unavailable rather than silently changing its denominator.
Both $L_i^{\mathrm{ref}}$ and $L_i$ measure planar base motion
only; interaction duration, attempt count, and interaction cost are excluded
from SPL and handled by IP and Total Cost.

\paragraph{ISR.}
Let $R_{ip}$ be the required interaction IDs in valid plan $p$ for episode
$i$, and $G_i$ the IDs with verified required effects. On required episodes,
\[
\mathrm{ISR}_i=\max_{p:\,R_{ip}\ne\emptyset}
\frac{|G_i\cap R_{ip}|}{|R_{ip}|},
\]
and
\[
\mathrm{ISR}=\frac{1}{|\mathcal{D}_{\mathrm{required}}|}
\sum_{i\in\mathcal{D}_{\mathrm{required}}}\mathrm{ISR}_i,
\]
with unnecessary episodes excluded. A required episode with no valid nonempty
plan makes aggregate ISR unavailable. A required ID earns credit at most once
per episode, including when its verified local effect occurs inside a failed
macro. Opening requires the terminal $0.8$ threshold or a verified transient
opening. ISR does not independently test passage or target visibility after
each effect; complete-plan success is a separate binary diagnostic. Thus a
mixed episode may receive partial ISR even if the full chain and final
navigation goal fail.

\paragraph{Target-Class IP.}
For episode $i$, let $\mathcal C_i$ denote the set of annotated interaction
classes, each specified by an interaction domain (channel or container) and an
object category. If $A_i$ is the number of interaction attempts and
$V_i^{\mathrm{class}}$ is the number of attempts producing at least one verified
physical effect on an object whose class is in $\mathcal C_i$, let $U_i=1$ only if
the benchmark declares interaction unnecessary, and $U_i=0$ otherwise. We define
\begin{equation}
\mathrm{IP}_i =
\begin{cases}
V_i^{\mathrm{class}}/A_i, & A_i>0,\\
U_i, & A_i=0,
\end{cases}
\qquad
\mathrm{IP}=\frac{1}{N}\sum_{i=1}^{N}\mathrm{IP}_i.
\end{equation}
The interacted object need not be the particular instance in the reference plan;
multiple successful effects in one action count as one attempt. An open effect
must cross from below $0.8$ to at least $0.8$ during the attempt, including
verified transient drawer opening. Reopening after closure can earn credit
again; an effect-free repeated request cannot. A partially successful macro
can earn one IP credit even if its overall acknowledgement reports failure.
Interactions
with other classes remain in the denominator, reflecting exploration under
incomplete information, but do not count as errors solely because they are
outside $\mathcal C_i$.

\paragraph{Total Cost.}
The operation cost and normalized episode score are
\begin{equation}
C_i^{\mathrm{op}}=L_i+\lambda A_i+\mu E_i,
\qquad
\mathrm{Cost}_i=\begin{cases}
\min(C_i^{\mathrm{op}}/B,1),&S_i=1,\\
1,&S_i=0,
\end{cases}
\label{eq:app-episode-cost}
\end{equation}
and $\mathrm{TotalCost}=N^{-1}\sum_i\mathrm{Cost}_i$. The default protocol
freezes $\lambda=0.3$, $\mu=1$, and $B=30$ before evaluation. $B$ is a
shared weighted operation-cost ceiling, not a rollout stopping rule. Failures
and successes over budget both score one. Successful effects outside target
classes pay the ordinary attempt cost without the additional error charge.
The unnormalized operation cost remains available as a diagnostic. Results
with different metric definitions, budgets, or weights are not pooled.

\paragraph{Aggregation.}
Each method's primary row averages scoring-eligible completed episodes; indented
rows use the channel, container, and mixed subsets. Overall is not an unweighted
average of split means. IP first computes a ratio within each episode and then
averages those ratios, rather than pooling attempts. ISR includes required
episodes only.

\paragraph{Split-level full-model diagnostics.}
Ours reaches
60.09\% ISR on required Channel episodes and 46.70\% on Container episodes.
Container IP is 38.91\%, whereas Mixed IP is 47.16\% despite 24.10\%
navigation SR. These quantities measure
different outcomes: ISR credits completed required effects, IP credits
target-class effects among attempted interactions, and SR requires the final
distance and visibility conditions. Mixed Cost is 0.9493: 759/1{,}000 Mixed
episodes fail navigation, and the metric assigns a cost of one to failures.

The main results table reports scores for 3{,}000 Ours episodes,
with 1{,}000 per split. ISR uses 2{,}654 required episodes (654 Channel and
1{,}000 each Container and Mixed). The MoMa-LLM adaptation also reports
scores for 3{,}000 scoring-eligible episodes, 1{,}000 per split, with the
same ISR denominators and $B=30$ cost ceiling. Its Channel, Container, and
Mixed SRs are 19.60\%, 10.30\%, and 1.70\%; Mixed ISR is 25.37\% and
Mixed IP is 39.81\%, illustrating the distinction between partial physical
effects and terminal navigation success.

\paragraph{Ablation evaluation.}
The ablation study compares Full with Flat Object Memory, Greedy Selection,
and Perception-only Update on the same fixed 200-episode Mixed subset
(Section~\ref{app:ablation_sampling}). Full is the reference within this evaluation,
with 23.0\% SR, 0.107 SPL, 41.8\% ISR, 40.4\% IP, and 0.961 Cost;
it is reported separately from the main benchmark aggregate.
Flat Object Memory has the highest ISR (43.8\%), while Greedy Selection
has the highest SPL (0.134) and lowest Cost (0.933). Full has the highest
SR and IP. Perception-only Update yields 5.5\% SR, highlighting the
largest observed completion gap among these variants.

Target visibility gain after opening a container is an optional diagnostic, not a
primary construction field.

\section{Additional Implementation Notes}
\label{app:impl}

\paragraph{Graph observations and attributes.}
Both MultINav-Agent and the MoMa-LLM adaptation use the same front-end
perception and the same restricted-GT observation interface in the primary
benchmark comparison. The interface provides episode-local IDs, semantic
categories, spatial observations, 2D masks or boxes, and 3D boxes for visible
entities; it does not expose hidden entities, joint metadata, or oracle plans.
The benchmark also supports an image-based raw-perception mode, but that mode
is not mixed with the restricted-GT comparison. Instance labels, segmentations,
and bounding boxes initialize and associate observed entities. Interactive
nodes store the interaction class, coarse state, candidate operation,
operable-part description, confidence, and observation recency. Visual
attribute inference refines these records from the current view. The coarse
candidate effects distinguish channel traversal from container-content
revelation; they do not establish the presence of an unobserved target.

\paragraph{Shared simulation interaction executor.}
Both interactive methods use the same executor in the primary evaluation.
The evaluator applies force-driven transitions to the selected articulated
joints, with a kinematic completion fallback when a transition does not reach
its requested target. Execution feedback is returned through the public
interaction interface; simulator joint metadata are withheld from the policies.
Agent graph updates maintain scene estimates, while the terminal evaluator
independently checks goal distance and head-camera visibility.

\paragraph{Scope of aggregate comparisons.}
Aggregate benchmark comparisons characterize end-to-end behavior; by themselves,
they do not isolate graph updates as the cause of a performance difference.

\paragraph{MoMa-LLM observation and decision interface.}
The reported MolmoSpaces adaptation receives head-camera RGB-D and restricted
simulator-derived observations of currently visible objects. Each restricted
record contains an episode-local opaque instance ID, a semantic category, a
2D bounding box and run-length mask, and a 3D axis-aligned bounding box
(center, size, and frame ID). Simulator object names, joint identities and
states, and oracle interaction plans are withheld from the policy. The adapter
uses these observations and depth to maintain an occupancy map and a dynamic
room-object graph. At high-level decision points, vLLM-served
Qwen3.6-35B-A3B-FP8 receives a summary of known rooms, objects, frontiers,
closed doors, and recent actions, then selects among
\texttt{navigate(room\_name, object\_name)},
\texttt{go\_to\_and\_open(room\_name, object\_name)},
\texttt{explore(room\_name)}, and \texttt{done()}.
The adapter converts the selected high-level action to navigation steps or a
public opening request; the evaluator executes the physical interaction and
verifies the terminal navigation goal. The model is queried at high-level
decision points rather than at every simulation step.
Its prompt summarizes the current room, nearby and previously discovered
object categories, unexplored frontiers, closed doors, and recent action
outcomes. Local navigation expands each selected high-level action into
waypoints; an opening request is sent through the benchmark's public
interaction interface, and its returned result updates the adapter's
object-state belief.

Full rollouts use the same episode identifiers and success criteria as their
light counterparts and may be recorded under different navigation/interaction
policies (default force joint control), allowing trajectory data to be joined
with the frozen evaluator manifest without changing benchmark membership.

\paragraph{Instruction disclosure.}
InstructionGoal supports hidden (goal only), partial (interaction may be needed),
and explicit (ordered objects/actions) language as an experimental factor over the
same grounded oracle plan.

\section{Vision-Language Model Decision Protocol and Prompts}
\label{app:vlm-protocol}

This section details the vision-language model (VLM) architecture, input representation, prompt templates, structured output schema, and decision context employed by MultINav-Agent.

\subsection{Model Configuration and Serving}
\label{app:vlm-config}

MultINav-Agent uses \texttt{Qwen3.6-35B-A3B-FP8} for high-level decisions,
served through an OpenAI-compatible chat-completions API.

Visual inputs are formatted as a composite RGB image containing:
\begin{itemize}
    \item \textbf{Global egocentric view}: The full head-camera observation with the candidate interaction target bounded by an explicit visual outline, providing geometric and spatial context relative to the robot and room.
    \item \textbf{Actionable region inset}: A padded, normalized crop around the target entity ($x, y \in [0, 1]$), allowing fine-grained localization of actionable door handles, push panels, or drawer pull surfaces.
\end{itemize}

\subsection{Subgoal Categorical Reason Codes and Subgraph Context}
\label{app:vlm-subgoal}

When evaluating candidate actions from the state-conditioned graph, the VLM receives serialized topological context containing:
\begin{enumerate}
    \item \textbf{Room and Anchor Entities}: Discovered room partitions and up to 4 major anchor objects per room (e.g., \texttt{refrigerator}, \texttt{countertop}, \texttt{bed}, \texttt{table}).
    \item \textbf{Candidate Portals and Containers}: Articulated doors and containers, their current states ($z_t$), and their mediated relational effects.
    \item \textbf{Decision Reason Codes}: The VLM classifies the semantic justification for each candidate into standardized reason codes:
    \begin{itemize}
        \item \texttt{TARGET\_VISIBLE}: The goal object is observed in line-of-sight; navigate to terminal observation pose.
        \item \texttt{UNLOCK\_ROUTE}: Opening a door is necessary to restore traversal connectivity to an unvisited topological room.
        \item \texttt{REVEAL\_TARGET\_CONTAINER}: Opening a container (e.g., refrigerator or cabinet) to expose a hidden target object matching semantic category priors.
        \item \texttt{EXPLORE\_TARGET\_ROOM}: Navigating to a newly connected or unexplored room when direct target evidence is absent.
    \end{itemize}
\end{enumerate}
This structured protocol makes candidate choices inspectable; the selected
action still requires execution and outcome verification.

\subsection{High-Level Candidate Selection Prompt}
\label{app:vlm-selection-prompt}

At each high-level decision point, the model receives the target request,
observed graph, current executable candidates, and recent action outcomes.
The selection instruction asks it to distinguish observations from semantic
priors, respect route prerequisites, avoid repeated failed actions, and return
only IDs from the current candidate list. The compact prompt form is:

\begin{quote}
\small\ttfamily\raggedright
Find the requested target. Rank up to three CURRENT executable candidate IDs.
Prefer an observed target goal; otherwise account for post-interaction traversal,
route prerequisites, plausible target containers, useful frontiers, and recent
failures. Do not infer hidden target locations or invent candidates. Return only
\{"ranked\_ids":[...],"reason":"...",\\
"confidence":"low|medium|high"\}.
\end{quote}

The response is checked against the current executable candidate IDs before
the chosen action is sent to navigation or interaction execution.

\end{document}